# Driving Decision and Control for Automated Lane Change Behavior based on Deep Reinforcement Learning

Tianyu Shi[1], Pin Wang[2], Xuxin Cheng[1], Ching-Yao Chan[2], Ding Huang[3]

*Abstract*— **To fulfill high-level automation, an automated vehicle needs to learn to make decisions and control its movement under complex scenarios. Due to the uncertainty and complexity of the driving environment, most classical rule-based methods cannot solve the problem of complicated decision tasks. Deep reinforcement learning has demonstrated impressive achievements in many fields such as playing games and robotics. However, a direct application of reinforcement learning algorithm for automated driving still face challenges in handling complex driving tasks. In this paper, we proposed a hierarchical reinforcement learning based architecture for decision making and control of lane changing situations. We divided the decision and control process into two correlated processes: 1) when to conduct lane change maneuver and 2) how to conduct the maneuver. To be specific, we first apply Deep Q-network (DQN) to decide when to conduct the maneuver based on the consideration of safety. Subsequently, we design a Deep Q-learning framework with quadratic approximator for deciding how to complete the maneuver in longitudinal direction (e.g. adjust to the selected gap or just follow the preceding vehicle). Finally, a polynomial lane change trajectory is generated and Pure Pursuit Control is implemented for path tracking for the lane change situation. We demonstrate the effectiveness of this framework in simulation, from both the decision-making and control layers.**

## I. INTRODUCTION

As automated driving systems become closer to deployment, the issues of safety and robustness of these systems operating in the real world are drawing greater attention. To fulfill SAE level 4 or 5 automation, the vehicle needs to learn when and how to make the right decision as well as how to execute the action safely. Especially when the vehicle is in an interactive environment, such as a lane changing scenario, the action of the surrounding vehicles may be highly unpredictable. A study shows that nearly 10 percent of all highway crashes are caused by lane change maneuvers [1]. Therefore, a safe, smooth and efficient lane change maneuver is an essential task for automated vehicles. To realize this function, the vehicle architecture should allow efficient and robust execution to deal with the uncertainties in the operating environment, and to make proper decision and perform reasonable action in response to the potentially adversarial or cooperative actions exhibited by the surrounding vehicles.

A considerable body of literature applied pre-defined rule-based model to address the path-planning and trajectory tracking problems under interactive situations. For example, the approaches of a potential field model and model predictive control are suggested in works by Rasekhipour, Y. et al. [2], Kim, B. et al. [3], and Ji, J., Khajepour. et al. [4]. However, as in a real-world scenario, some irrational and unforeseen behaviors (e.g. suddenly overtaking) may render the aforementioned methods inefficient.

Machine learning methods have demonstrated their capacity for solving complexed problems without rigid programming rules. Zhu, Y. et al. [5] demonstrated the good performance of a model-free deep reinforcement learning method that leverages a small amount of demonstration to assist a reinforcement learning agent. However, without a well-trained model and proper policies design, the behavior of the agent may still be unsatisfactory.

In this paper, we combine the classical control and machine learning methods in a hierarchical way by leveraging their advantages. We first use a DQN to decide when (i.e., now or not now) to perform the lane change, and then use a unique Q-learning approach with a quadratic Q-function to handle the challenge of the continuous control action in longitudinal direction (e.g. adjust for target gap or keep car-following). The first layer in our proposed framework will then to decide whether it is safe to make a lane change right now. If so, the second layer, integrated with classic control modules (e.g. Pure Pursuit Control) will output appropriate control actions for adjusting its position. Otherwise, it will follow the preceding vehicle on its own lane with the longitudinal control output by the Quadratic Q-function approximator. The second layer can be optimized and executed at each time step while the first layer can be performed at a relatively low frequency to lower the overall computation cost. In such a way, it can serve for the control demand of a high update frequency, and at the same time maintain the prompt response under unforeseen and dangerous situations. The main contribution in

Resrach supported by PATH, UC Berkeley.

1.Tianyu Shi and Xuxin Cheng are with Beijing Institute of Technology, Zhongguancun Street, Haidian District, Beijing, 100081, {tianyu.s,chengxuxindavid}@outlook.com

2.Pin Wang (corresponding author) and Ching-Yao Chan are with California PATH, University of California, Berkeley, 1357 South 46th Street., Richmond, CA. 94804, US. {pin_wang, cychan} @berkeley.edu

3. Ding Huang is with Chinese Academy of Launch Vehicle Techonology

our work is that we decompose the decision problem into two hierarchical layers where the decision layer supervises the driving environment and guarantees safety and the control layer provides smooth maneuver transition. Unlike some previous works (e.g. [9-12]), our agent has more autonomy by interweaving the decision and control in a hierarchical way. Our results show this design provide a reasonable and effective approach for the task of lane change.

## I. Related Work

Many studies have been carried out in the domain of decision making and control for automated vehicles. Some researchers consider it as an optimal control problem and apply predefined functions for optimization. For example, Nilsson, J., et al. [6] presented a lane change maneuver algorithm which could determine whether a lane change was desirable and if an inter-vehicle gap and time instance to perform was appropriate based on the predefined utility function. Luo, Y., et al. [7], proposed a dynamic lane change maneuver where they converted the planning problem into a constrained optimization problem using the lane change time and distance. These aforementioned algorithms can define a smooth path for lane changing theoretically, however, they are not robust enough in some undefined situations. Also, they cannot formalize a robust policy to the traffic environment subject to uncertainty and sensor noise.

Machine learning approach have been applied to various problems for dealing with unforeseen situations, on the condition that the algorithms are properly trained on a large set of sample data without explicit boundaries or control rules. Vallon, C, et al. [8] used Support Vector Machine (SVM), a supervised learning method, to predict the lane-change-initiation behavior for automated driving based on personalized human driver data. However, if the labeled data is noisy or the human driver in the experiment shows irregular behaviors, the learned results may be unreliable. The trained behavior will also be more personalized rather than common to all drivers. Some researchers explored imitation learning for driver behavior cloning to generate driver-like trajectories. Bhattacharyya, et al. [9], extended Generative Adversarial Imitation Learning (GAIL) to address these shortcomings and used a Parameter Sharing approach to guarantee imitation learning in a multi-agent context. As a result, a new algorithm called PS-GAIL was generated. Although they addressed challenges of instability to some extent in the learning environment through a training curriculum, the emergent values (e.g. collision rate, hard brake rate) climbed up with the increase of number of agents.

Some researchers proposed the deep reinforcement learning framework to achieve a robust and reliable automated driving policy. Wang, P., et al. [10] proposed to apply Deep Reinforcement Learning (DRL) techniques for finding an optimal driving policy by maximizing the long-term reward in an interactive environment. Wu, C et al. [11] integrated a traffic simulator with DRL library to develop reliable policy for complex multi-agent problems, such as mixed autonomy traffic (involving both automated and human driving vehicles). However, both of the above works used a rule-based car following model to simplify the longitudinal control. In fact, this is a drawback which cannot help the vehicle to learn a more naturalistic behavior. In one previous work [12], a reinforcement learning based approach was presented for automated lane change maneuver, which allows the agent to explore the unforeseen environment and make the correct decision based on the Q-values. However, the ego vehicle's lane change process is without in-depth exploration of decision making.

## II. Problem Description

In a typical lane change scenario as shown in Fig 1, the ego vehicle must be able to adjust its actions to fit into the dynamic traffic environment. Therefore, the feasible lane change policy $\pi$ to be learned should be able to:

- Avoid collision with surrounding vehicles,
- Conduct smooth movement.

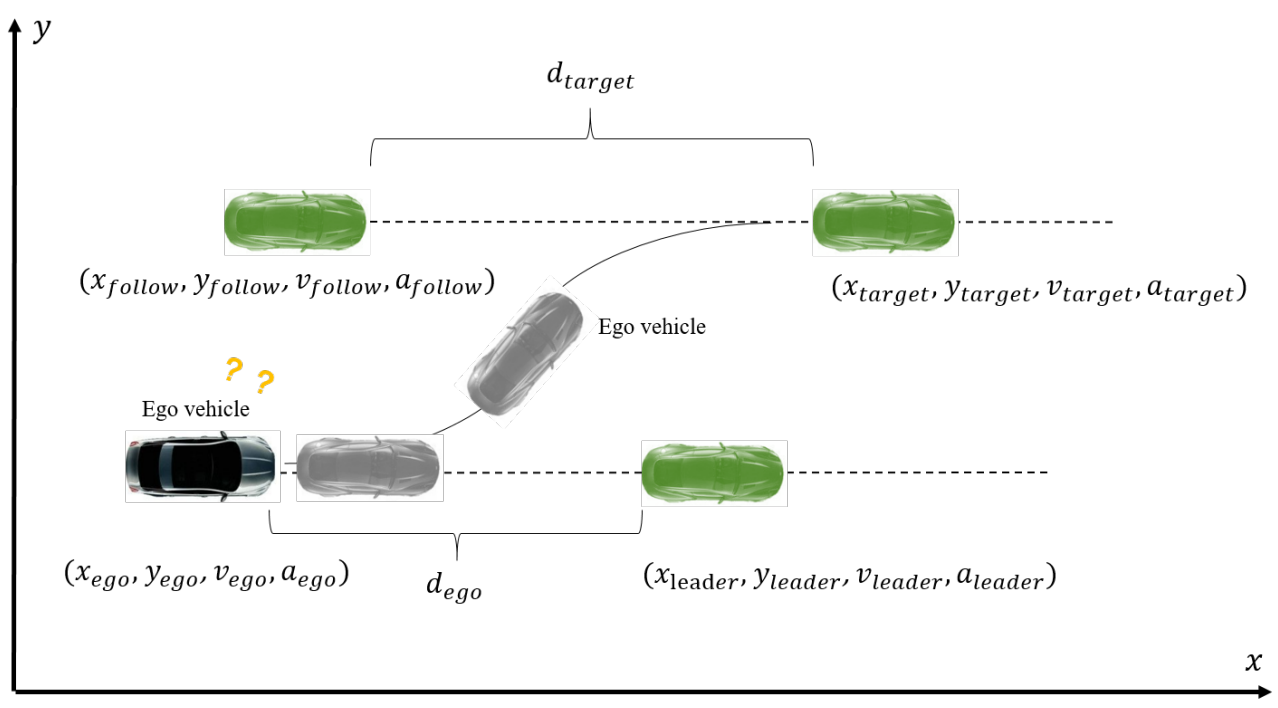

Figure 1. Example of a lane change scenario

Generally speaking, the decision making includes strategic and tactical strategies. The strategic decision means that the lane change will be initiated by the longer-term goals such as travel efficiency (e.g. traffic flow conditions in multiple lanes), the remaining time in the current lane (e.g. a need to exit from a nearby ramp), and other relevant factors on route planning over the longer time horizon. The tactical decision-making means that the ego vehicle already have the intention for lane changing but it needs to follow through. Our focus in this paper goes to the tactical decision level.

For the first layer, the state space includes the relative distance between the surrounding vehicles ($\Delta x_{leader}$, $\Delta x_{target}$, $\Delta x_{follow}$), and the relative velocity between the ego vehicle and its leader $\Delta v_{leader}$ and that between the ego vehicle and the leader vehicle in the target lane $\Delta v_{target}$; the speed $v_{ego}$, and acceleration $a_{ego}$ of the ego vehicle.

$$s = (\Delta x_{leader}, \Delta x_{target}, \Delta x_{follow}, \Delta v_{leader}, \Delta v_{target}, v_{ego}) \in S \tag{1}$$

The action space is defined as follows as evaluating whether the lane change movement is beneficial.

$$a_I = \{0,1\} \in A_I \tag{2}$$

where '0' means 'no lane change right now', i.e., remaining in the ego lane, and '1' represents 'lane change right now'.

For the second control layer, the task is to consider how to conduct the maneuver appropriately. In this task, we divide it into two steps, adjustment step and execution step. The 'adjustment step' is to adjust the longitudinal action for a suitable opportunity. Meanwhile, the 'execution step' is to perform smooth and effective movement to complete the maneuver. To be specific, if the lane change decision is '1', i.e., make change lane at this moment, an action adjustment module will function and the ego vehicle will adapt its longitudinal action to fit into a gap on the target lane, and then in the execution step, a planned trajectory will be generated and a controller based on Pure Pursuit Control will track the reference trajectory. If the lane change decision is 0, i.e., not to change lane immediately, the adjustment step and execution step will be merged into one, and a car following module will function by which the ego vehicle will remain in the lane and wait for another gap.

In the action adjustment step, the action space is defined with continuous longitudinal acceleration $a_E$ as:

$$a_E = a_{ego} \in A_E \tag{3}$$

Reinforcement Learning is used to train the decision module and the longitudinal control module within the hierarchical structure to help our agent not only learn the high-level policy for deciding when to make lane changes but also learn the low-level control policy for adjusting its status to a feasible maneuvering condition. The architecture of our overall decision-making framework is shown in the Fig 2.

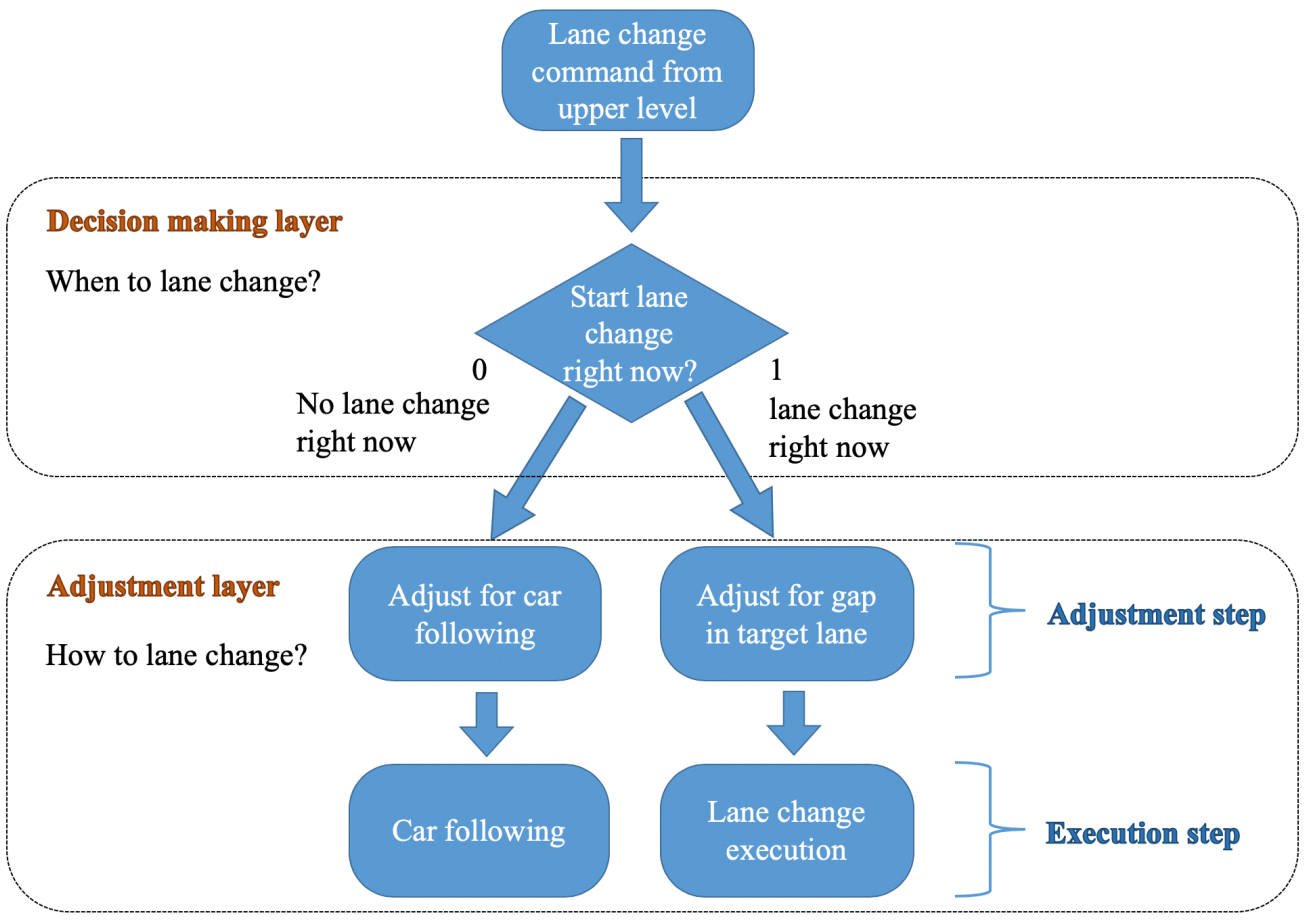

Figure 2. Structure of the proposed framework

## III. METHODOLOGY

### A. Decision making layer

In the decision-making layer, a discrete decision is given by a Q-network which is designed with two fully connected layers in our study. The reward function of the decision-making module is based on three factors, i.e., distance $\Delta x_{ego}$ between ego vehicle and leading vehicle, the relative speed $\Delta v_{ego}$ between ego vehicle and leading vehicle, and the target gap $d_{gap}$ on the target lane. For example, if the distance $\Delta x_{ego}$ is big enough, the relative speed $\Delta v_{ego}$ is small, and the gap $d_{gap}$ is narrow, the ego vehicle will tend to remain in the ego lane and keep following the front vehicle. On the other hand, if the relative distance $\Delta x_{ego}$ is small and the gap $d_{gap}$ is big enough, the ego vehicle will tend to change to the target lane.

Therefore, we can construct the reward function for the decision-making problem as in the following equation:

$$r_{safe} = \begin{cases} w_1\left|d_{ego} - x_{leader} - x_{ego}\right| + w_2\left|v_{ego} - v_{leader}\right|, a_I = 0 \\ w_3\left(d_{target} - d_{gap}\right) + w_4\left|v_{ego} - v_{target}\right|, a_I = 1 \end{cases} \tag{4}$$

where $d_{target}$ is the desired distance in the target lane and can be calculated based on (5), $d_{ego}$ is the desired distance in the target lane that in turn can be calculated based on (6) where $d_t$ is a safety threshold, $a$ is the maximum acceleration, $\tau$ is the reaction time of the human action, $d_0$ is the minimum distance and $t$ is the total time of the lane change.

$$d_{target} = v_{ego}t + \left(x_{target} - x_{ego}\right) + \tau(v_{target} - v_{ego}) + d_0 \tag{5}$$

$$d_{ego} = v_{ego}\tau + \frac{v_{ego}^2}{2a} + d_t \tag{6}$$

### B. Adjustment step

In the adjustment step, we need to calculate the continuous action for the maneuver which is output by the decision-making layer. However, the traditional Q-learning method cannot handle problems with continuous action space.

There are policy gradient-based algorithms, e.g. actor-critic, to directly learn the policy without resorting to a value function, but it still needs much effort on designing the policy network. In the adjustment layer, we use a modified Q-function approximator for handling the continuous action space, which is similar to our previous work [12]. To be specific, we design a Q-function in a quadratic form so that the optimal action has a closed form solution. Our method can also simplify the network design effort in the learning algorithm.

The Q-function approximator is expressed as follows:

$$Q(s,a) = A(s) \cdot (B(s) - a)^2 + C(s) \tag{7}$$

where $A$, $B$, and $C$ are coefficients and designed with neural networks with the state information as input, as illustrated in Fig 3. Because any smooth Q-function can be Taylor expanded in this form near the greedy action while there is not much loss in generality if we stay close to the greedy action in the Q-learning exploration.

As shown in Fig 3, $A$ is designed with one hidden layer of 150 neurons, and particularly $A$ is bounded to be negative with the use of a soft-plus activation function on the output layer. $C$ is also a one-hidden-layer neural network with the same number of neurons as in $A$, and its output can be any scalar number. B is a neural network that consists of two hidden layers with 200 neurons and ReLU activation function in each layer.

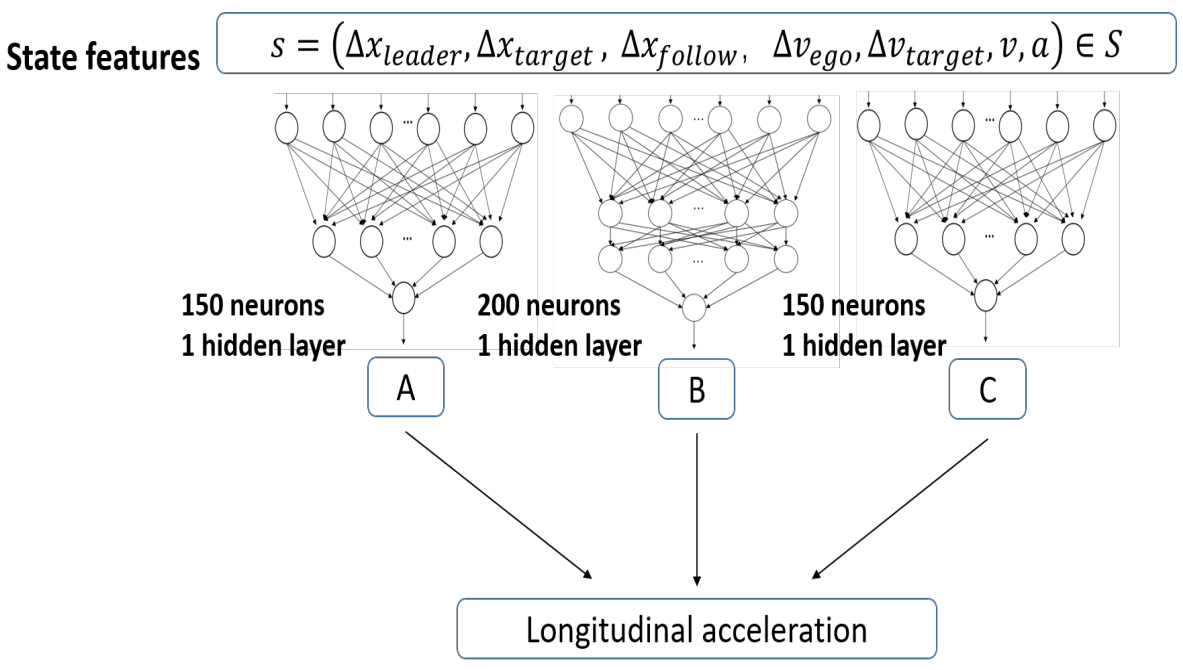

Figure 3. Structure of Q function approximator

The adjustment module for the two discrete decisions (e.g. 'lane change right now' and 'no lane change right now') are trained with the same Quadratic Q-network structure but with different reward functions and weights which we will discuss in details later.

*1) Reward for car following module*

In a car following situation, we need to consider the distance as well as the relative speed between ego vehicle and leader vehicle. A good car following module should maintain proper relative distance and speed while keeping the acceleration of ego vehicle within a comfortable range. Based on this idea, we designed the following reward function.

$$R_{dis} = -w_{dis} \cdot \left| x_{leader} - x_{ego} - d_{ego} \right| \tag{8}$$

where $R_{dis}$ is the reward related to relative distance, $w_{dis}$ is the weight of the reward, $x_{leader}$ is the longitudinal position of the leader vehicle, $x_{ego}$ is the longitudinal position of the ego vehicle, $d_{ego}$ is the desired distance to the leader vehicle.

$$R_{\Delta v} = -w_{\Delta v} \cdot \left| v_{ego} - v_{leader} \right| \tag{9}$$

$R_{\Delta v}$ is the reward related to relative speed, $-w_{\Delta v}$ is the weight of the reward, $v_{ego}$ and $v_{leader}$ are the speed of ego vehicle and leader vehicle.

$$R_C = R_{dis} + R_{\Delta v} \tag{10}$$

$R_C$ is the total reward for the car following module. It is the summation of $R_{\Delta v}$ and $R_{dis}$.

*2) Reward for lane change module*

In the lane change situation, the ego vehicle should adjust its longitudinal acceleration or deceleration $a_E$ in order to fit into the target gap while remaining safe distance to the leader vehicle in ego lane. The reward function is defined as:

$$r_{dis} = -w_{dis} \cdot \left| min(\Delta x_{leader}, \Delta x_{target}) - \Delta x_{follow} \right| \tag{11}$$

That means the ego vehicle's desired position is the middle between the follow vehicle and the closer between the leader vehicle and the target vehicle, as described in Fig. 1.

$$R_{\Delta v} = -w_{\Delta v} \cdot \left| v_{ego} - \min(v_{leader}, v_{target}) \right| \tag{12}$$

$R_{\Delta v}$ has the same meaning as in car following module.

$$R_A = R_{dis} + R_{\Delta v} \tag{13}$$

The total reward $R_A$ is the sum of $R_{dis}$ and $R_{\Delta v}$.

*C. Execution step*

The execution step is separated from the adjustment step for the lane change decision while it is merged with the adjustment step for the car-following decision. Therefore, in this subsection, we describe how to execute the lane change movement when the automated vehicle acquires a lane change signal from the decision layer.

A reference trajectory based on fifth degree polynomial curve will be generated with the initial states $(x_i, \dot{x}_i, \ddot{x}_i, y_i, \dot{y}_i, \ddot{y}_i)$ and the terminal states $(x_t, \dot{x}_t, \ddot{x}_t, y_t, \dot{y}_t, \ddot{y}_t)$. The format of the reference trajectory is written as in (14) and (15).

$$x(t) = a_5 t^5 + a_4 t^4 + a_3 t^3 + a_2 t^2 + a_1 t + a_0 \tag{14}$$

$$y(t) = b_5 t^5 + b_4 t^4 + b_3 t^3 + b_2 t^2 + b_1 t + b_0 \tag{15}$$

Then we can define the time parameter matrix:

$$T_{6x6} = \begin{bmatrix} t_i^5 & t_i^4 & t_i^3 & t_i^2 & t_i & 1 \\ 5t_i^4 & 4t_i^3 & 3t_i^2 & 2t_i & 1 & 0 \\ 20t_i^3 & 12t_i^2 & 6t_i^1 & 2 & 0 & 0 \\ t_i^5 & t_t^4 & t_i^3 & t_t^2 & t_t^1 & 1 \\ 5t_t^4 & 4t_t^3 & 3t_t^2 & 2t_t^1 & 1 & 0 \\ 20t_t^3 & 12t_t^2 & 6t_t^1 & 2 & 0 & 0 \end{bmatrix} \tag{16}$$

For the lane change motion control, we adopted the Pure Pursuit control method to follow the reference trajectory. By generating the target waypoints with velocity and position. We can control the ego vehicle to keep track of the target trajectory (as shown in Fig. 4). By calculating the angle α between the vehicle body and the target waypoint, we can control steering wheel angel to track the target point.

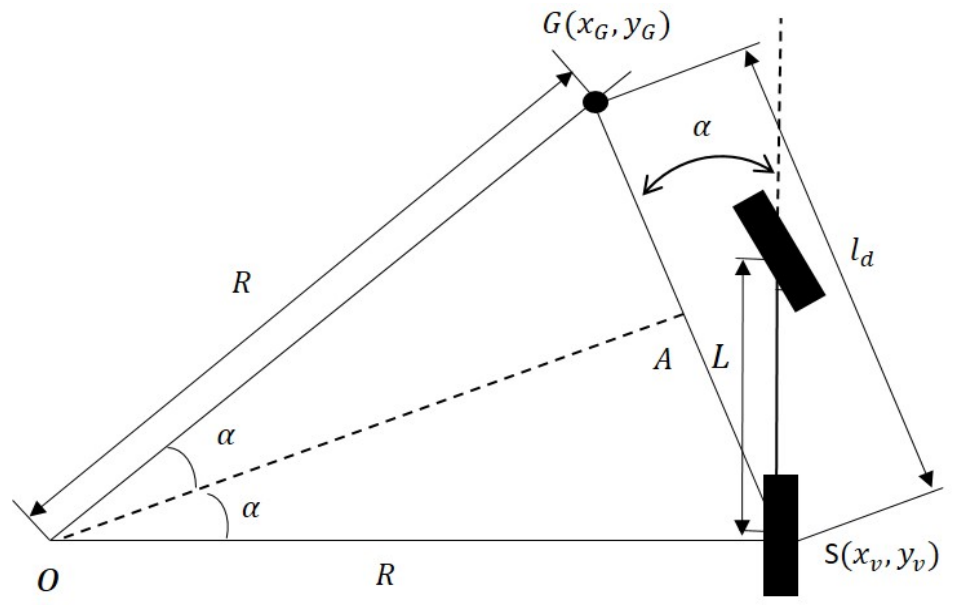

Figure 4. Path tracking based on Pure Pursuit Model

According to the geometry relationship, we can get:

$$\frac{l_d}{sin\alpha} = 2R \tag{17}$$

$$\delta(t) = tan^{-1}(\frac{2Lsin\big(\alpha(t)\big)}{l_d}) \tag{18}$$

where, $l_d$ is the look ahead distance. Larger $l_d$ will make the tracking smoother and smaller $l_d$ will make the tracking more accurate $\alpha$ is actual front steering wheel angle, and $\delta(t)$ is the controlled front steering wheel angle, which keeps track of the orientation. $L$ is the wheel base of the vehicle.

## IV. Implementation on Lane Change Scenario

### A. Simulation Scenario Building

We build a simulation environment where the ego vehicle is controlled with our proposed hierarchical RL framework and the other surrounding vehicles always travel on its lane and are controlled based on the Intelligent Driver Model (IDM) [12].

### B. Implementation on Lane change scenario

First, we build our simulation environment as depicted in Fig. 5. The red cars are ego vehicles, the green cars are vehicles associated with the selected gap, and the blue cars are normal vehicles controlled by IDM.

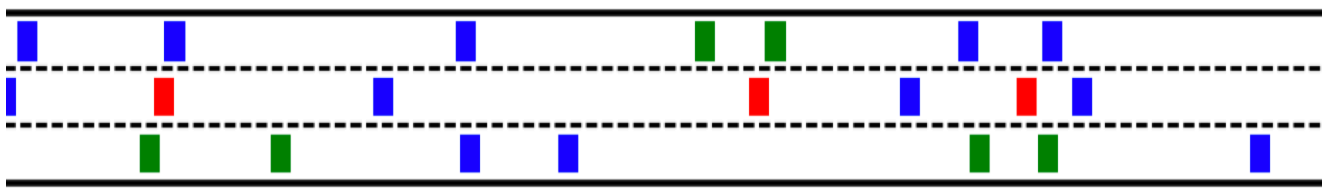

Figure 5. Simulation environment

In this simulation platform, the segment length is 1000m and each lane width is 3.75m. The initial speed of vehicles varies from 75km/h to 136km/h. After a certain amount of time the leader vehicle was generated and, the ego vehicle will be generated at the start of the lane. The period $t_e$ between the generation of ego vehicle and leader vehicle is set to $U(0.1, 7)$, representing a uniform distribution. Thus the initial distance between the leader vehicle and the ego vehicle is $\int_0^{t_e} v_{leader}\, dt\,, t_e \sim U(0.1, 7)$. These settings allow the environment to generate distinctive situations so that the behavior learned by agent is robust. In addition, as shown in the Fig 3, the initial values of A and B are set as 0, and that of C -60.

During the training process, we choose the learning rate as 0.0005, replay buffer size is 5000, and batch size is 64, and a decayed $\epsilon$ from 1 to 0.1 in 300,000 steps. At first, the agent tends to take random actions to explore the environment and acquire information of the environment. As the training process moves on, the agent is more likely to exploit learned rewarding mechanism to make the reasonable decisions.

The loss curve along training steps and accumulated reward per episode are depicted as Fig 6, from which we can see that the loss and rewards converge well, indicating that the agent successfully learns to take actions to maximize the reward. To better evaluate the performance of the model, we chose a typical initial condition where the initial speed of leader vehicle is 35.5 m/s while the initial speed of the ego vehicle is 17 m/s and the relative distance between them is 68 m.

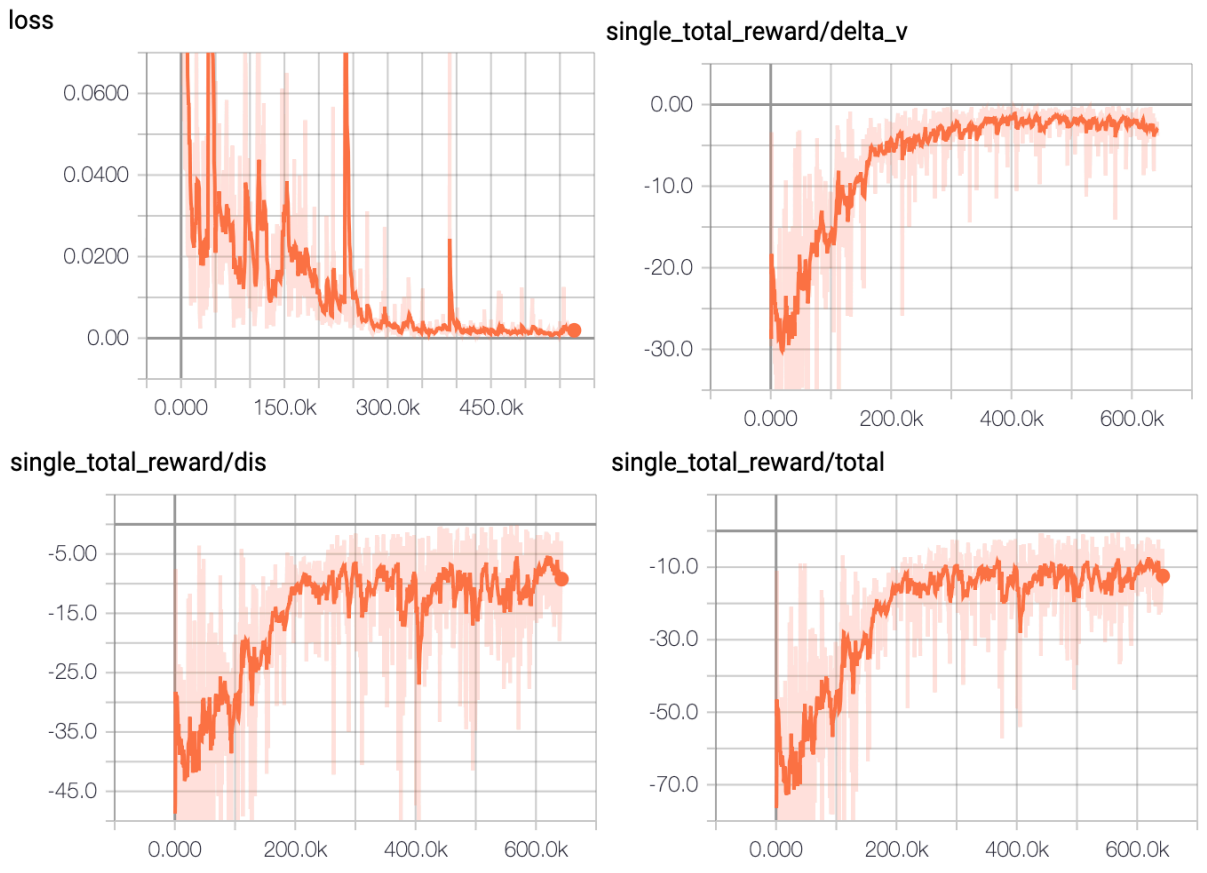

Figure 6. Evaluation of training loss and reward fuction

Fig 7 demonstrates that the ego vehicle successfully learns to adjust its speed and relative distance to a proper value after about 100 timesteps (10s). Even though there is a speed difference between the ego vehicle and leading vehicle at the start, we can see that the ego vehicle achieves good performance, reaching small relative speed.

We first train the longitudinal adjustment module under car-following and lane-changing situations, and then train the decision-making layer while holding the parameter in the controller layer fixed. From the simulation results, we can conclude that the proposed structure is able to solve the lane change problem with respect to the reward function designed.

From the results of the experiments, the decision-making module and the adjustment module both achieve good performance. The ego vehicle is able to recognize the safety of a gap when lane change command is committed.

Furthermore, we did evaluation on the lane change performance. The curve in Fig. 8 shows that the ego vehicle can successfully and smoothly change to the target lane without abrupt lateral acceleration or deceleration. Fig. 8(a) shows the longitudinal velocity and yaw acceleration during a lane change maneuver. Fig. 8(b) shows the trajectory with respect to the X and Y coordinates during a lane change maneuver.

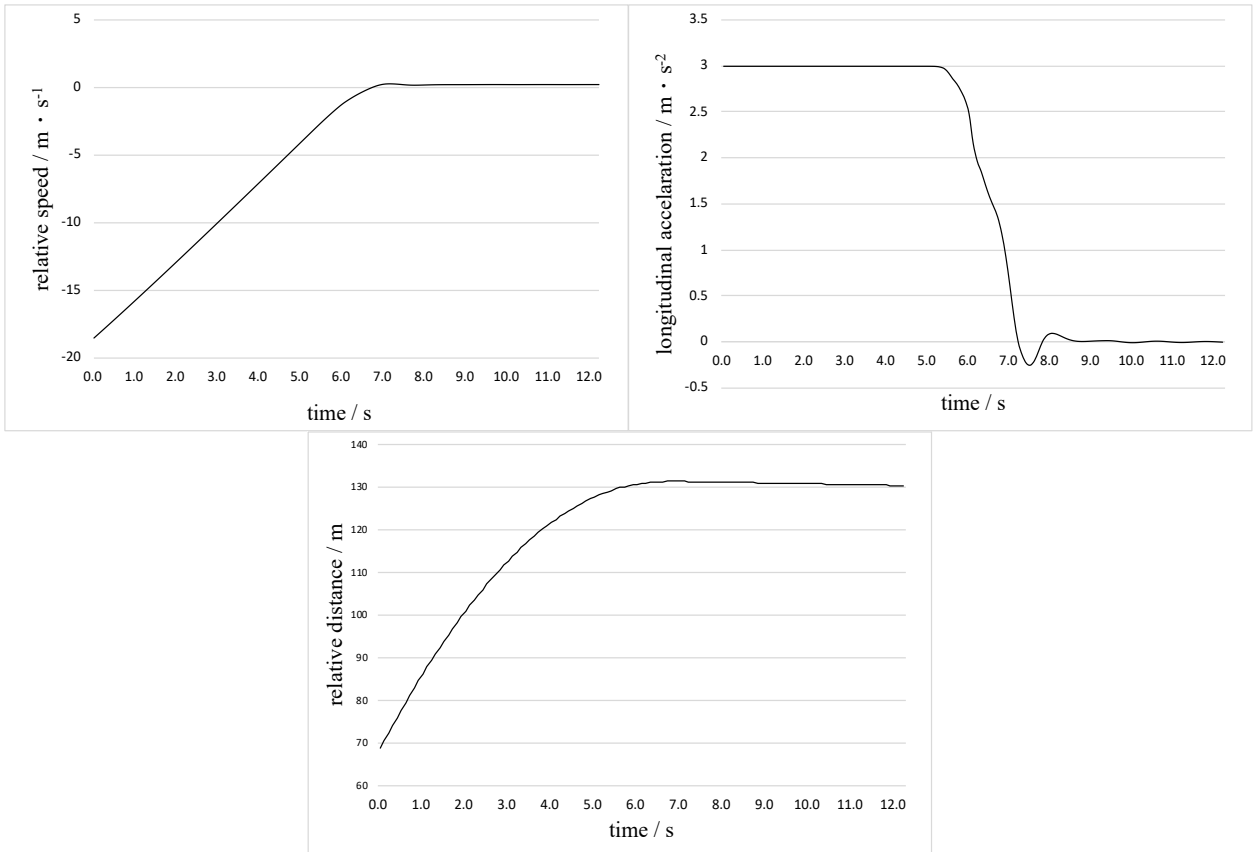

Figure 7. Evaluation of adjustment step

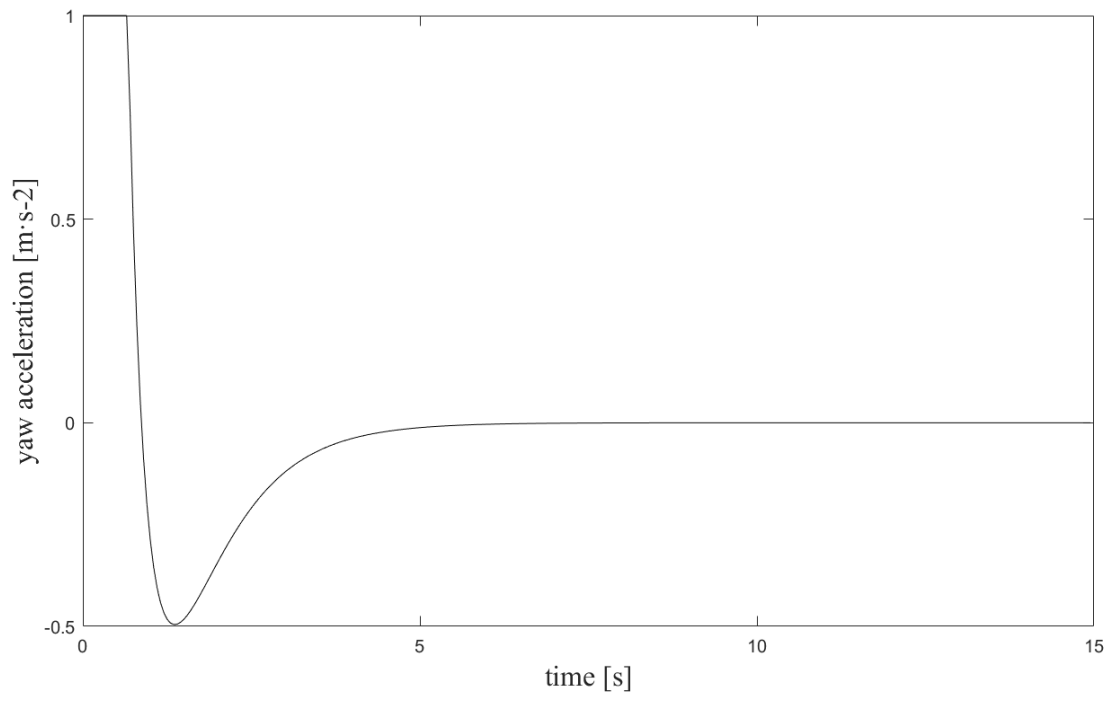

(a) Lane change yaw acceleration evaluation

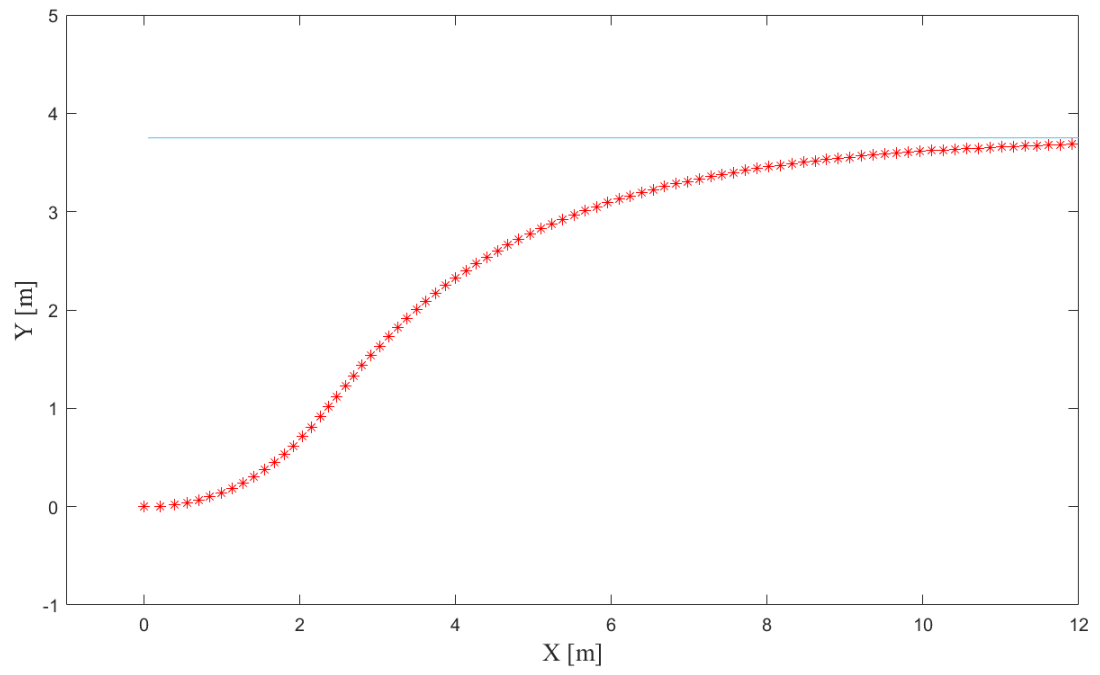

(b) lane change trajectory

Figure 8. lane change performance evaluation

## V. Conclusion And Discussion

For the problem of lane change, a simple deep reinforcement learning method is not able to complete the whole task due to the task's complexity. Our proposed structure of hierarchical DQN decomposes the entire procedure into the decision layer and the control layer where the control layer consist of an adjustment step and execution step. In the adjustment step, two longitudinal action adjustment modules are trained with the same quadratic Q-network framework but different reward functions, and then the decision layer is trained subsequently to learn an overall policy.

Training and testing results from simulation show the feasibility of the proposed approach. The loss curve and accumulated reward demonstrate converging trend, implying that the agent has learned to make lane changes or lane keeping under different situations. The kinematic curves further validate that the vehicle agent can execute the decision smoothly.

The next step of our research is to focus more on other types of decisions (e.g. overtaking) to make the decision making strategic more diverse and robust. Different reinforcement learning methods can also be applied to continuous action space (e.g. Deep Deterministic Policy Gradient) to improve performance. To better exploit the capability of reinforcement learning in discrete space, we can also use a more complicated hierarchical framework (e.g. Option Graph) to handle decision making, and give control and execution with classical control methods (e.g. MPC). The suggested approach potentially can be adapted to various situations other than lane change. The simulation environment also has much potential to be improved. The state-of-the-art traffic simulation environment such as SUMO [13] is also worth trying.

## References

[1] S. Hetrick. "Examination of driver lane change behavior and the potential effectiveness of warning onset rules for lane change or "side" crash avoidance systems," Dissertation, Virginia Polytechnic Institute & State University, 1997.

[2] Y. Rasekhipour, A. Khajepour, S.-K. Chen, and B. Litkouhi, "A Potential Field-Based Model Predictive Path-Planning Controller for Autonomous Road Vehicles," *IEEE Trans. Intell. Transp. Syst.*, vol. 18, no. 5, pp. 1255–1267, May. 2017.

[3] B. Kim, D. Necsulescu, and J. Sasiadek, "Model predictive control of an autonomous vehicle," in *Proc. IEEE/ASME Conf. Advanced Intell. Mechatronics.,* 2001, vol. 2, pp. 1279–1284.

[4] J. Ji, A. Khajepour, W. W. Melek, and Y. Huang, "Path Planning and Tracking for Vehicle Collision Avoidance Based on Model Predictive Control With Multiconstraints," *IEEE Trans. Veh. Technol.*, vol. 66, no. 2, pp. 952–964, Feb. 2017.

[5] Y. Zhu et al., "Reinforcement and Imitation Learning for Diverse Visuomotor Skills," ArXiv180209564 Cs, Feb. 2018.

[6] . J. Nilsson, J. Silvlin, M. Brannstrom, E. Coelingh, and J. Fredriksson, "If, When, and How to Perform Lane Change Maneuvers on Highways," *IEEE Intell. Transp. Syst. Mag.*, vol. 8, no. 4, 2016, pp. 68–78.

[7] . Y. Luo, Y. Xiang, K. Cao, and K. Li, "A dynamic automated lane change maneuver based on vehicle-to-vehicle communication," *Transp. Res. Part C Emerg. Technol.*, vol. 62, pp. 87–102, Jan. 2016.

[8] .C. Vallon, Z. Ercan, A. Carvalho, and F. Borrelli, "A machine learning approach for personalized autonomous lane change initiation and control," in *Proc. IEEE Intell. Veh. Symp*., 2017, pp. 1590–1595.

[9] R. P. Bhattacharyya, D. J. Phillips, B. Wulfe, J. Morton, A. Kuefler, and M. J. Kochenderfer, "Multi-Agent Imitation Learning for Driving Simulation," in *Proc. IEEE/RSJ Int. Conf. Intell. Robot. Syst*., 2018, pp. 1534–1539.

[10] P. Wang and C. Chan, "Formulation of deep reinforcement learning architecture toward autonomous driving for on-ramp merge," in *Proc. IEEE Conf. Intell. Transp. Syst*., 2017, pp. 1–6.

[11] C. Wu, A. Kreidieh, K. Parvate, E. Vinitsky, and A. M. Bayen, "Flow: Architecture and Benchmarking for Reinforcement Learning in Traffic Control," ArXiv171005465 Cs, Oct. 2017.

[12] P. Wang, C. Chan, and A. d L. Fortelle, "A Reinforcement Learning Based Approach for Automated Lane Change Maneuvers," in Proc. IEEE Intell. Veh. Symp., 2018, pp. 1379–138

[13] Behrisch, M., Bieker, L., Erdmann, J., and Krajzewicz, D. , "SUMO–simulation of urban mobility: an overview,". In *Proc. SIMUL* 2011